\documentclass[runningheads]{llncs}
\usepackage{graphicx}
\usepackage{etoolbox}

\usepackage[mobile]{eccv}
\usepackage{eccvabbrv}

\usepackage{amsmath,amssymb}
\usepackage{booktabs}
\usepackage{multirow}
\usepackage{tcolorbox}
\tcbuselibrary{skins,breakable}
\usepackage{xcolor}
\usepackage{enumitem}
\usepackage{subcaption}
\usepackage{wrapfig}
\usepackage{colortbl}
\usepackage{adjustbox}
\usepackage{pifont}
\usepackage{tikz}

\usepackage{float}
\usepackage{fontawesome5}

\usepackage{array}
\usepackage{makecell}

\usepackage{xspace}

\definecolor{checkgreen}{HTML}{2E7D32}
\definecolor{crossred}{HTML}{BDBDBD}
\definecolor{partialyellow}{HTML}{E65100}
\definecolor{oursrow}{HTML}{E3F2FD}
\definecolor{sectiongray}{HTML}{757575}

\definecolor{checkgreen}{HTML}{2E7D32}
\definecolor{partialyellow2}{HTML}{F57F17}
\definecolor{crossred}{HTML}{C62828}
\definecolor{headerblue}{HTML}{1A237E}
\definecolor{headerbg}{HTML}{E8EAF6}
\definecolor{sportscol}{HTML}{E8F5E9}
\definecolor{rowalt}{HTML}{F5F5F5}

\newcommand{\cmark}{\textcolor{checkgreen}{\ding{51}}}
\newcommand{\xmark}{\textcolor{crossred}{\ding{55}}}

\newcommand{\pmark}{\textcolor{partialyellow}{$\boldsymbol{\sim}$}}

\definecolor{pillar1}{HTML}{4A7FB5}
\definecolor{pillar2}{HTML}{2E9E8F}
\definecolor{pillar3}{HTML}{D4913A}
\definecolor{pillar4}{HTML}{C0392B}
\definecolor{lightgray}{HTML}{F5F5F5}

\newtcolorbox{examplebox}[1][]{
  colback=gray!5,
  colframe=gray!50,
  fonttitle=\bfseries,
  title=#1,
  boxrule=0.5pt,
  arc=2pt,
}

\usepackage[accsupp]{axessibility}

\usepackage[hidelinks]{hyperref}
\newcommand{\name}{\textsc{SVI-Bench}\xspace}

\newcommand{\mypara}[1]{\vspace{4pt}\noindent\textit{#1}\enspace}

\makeatletter
\renewcommand\subsubsection{\@startsection{subsubsection}{3}{\z@}
  {-1.5ex plus -0.5ex minus -.2ex}{0.3ex plus .2ex}
  {\normalfont\normalsize\bfseries}}
\makeatother

\definecolor{pillarblue}{HTML}{4A7FB5}
\definecolor{pillarteal}{HTML}{2E9E8F}
\definecolor{pillaramber}{HTML}{D4913A}
\definecolor{pillarcrimson}{HTML}{C0392B}

\newtcolorbox{finding}[2]{
  colback=#1!6!white,
  colframe=#1!25!white,
  boxrule=0.4pt,
  arc=1.5mm,
  left=5pt, right=5pt,
  top=3pt, bottom=3pt,
  before skip=6pt,
  after skip=2pt,
  fontupper=\small,
  before upper={\textcolor{#1!80!black}{\textbf{#2}}\enspace}
}

\begin{document}



\DeclareRobustCommand{\svilogo}{%
  \raisebox{-0.15em}{\includegraphics[height=1.1em]{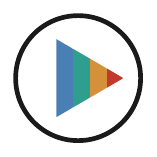}}%
}

\title{\texorpdfstring{\svilogo\hspace{0.2em}\textsc{SVI-Bench}: A Dynamic Microworld \\ for Strategic Video Intelligence}{SVI-Bench: A Dynamic Microworld for Strategic Video Intelligence}}

\author{Yulu Pan\inst{1}$^\star$ \and
Han Yi\inst{1}$^\star$ \and
Seongsu Ha\inst{1}$^\star$ \and
Md Mohaiminul Islam\inst{1}$^\star$ \and \\
Benjamin Zhang\inst{1} \and
Lorenzo Torresani\inst{2} \and
Gedas Bertasius\inst{1}}

\institute{
\raisebox{-0.3\height}{\includegraphics[height=2.0em]{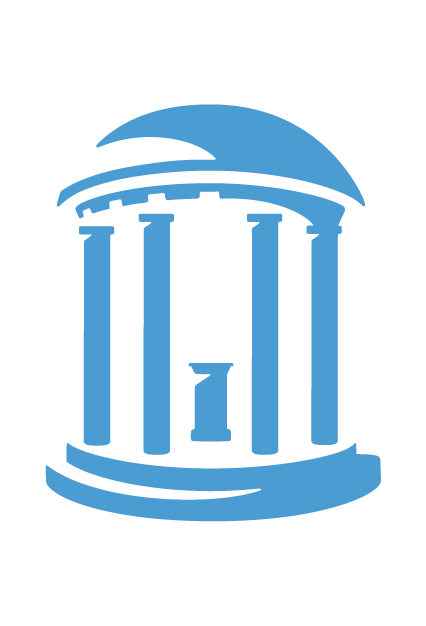}}\;
$^1$University of North Carolina at Chapel Hill, Chapel Hill, NC, USA
\\[0.2em]
\raisebox{-0.3\height}{\includegraphics[height=1.6em]{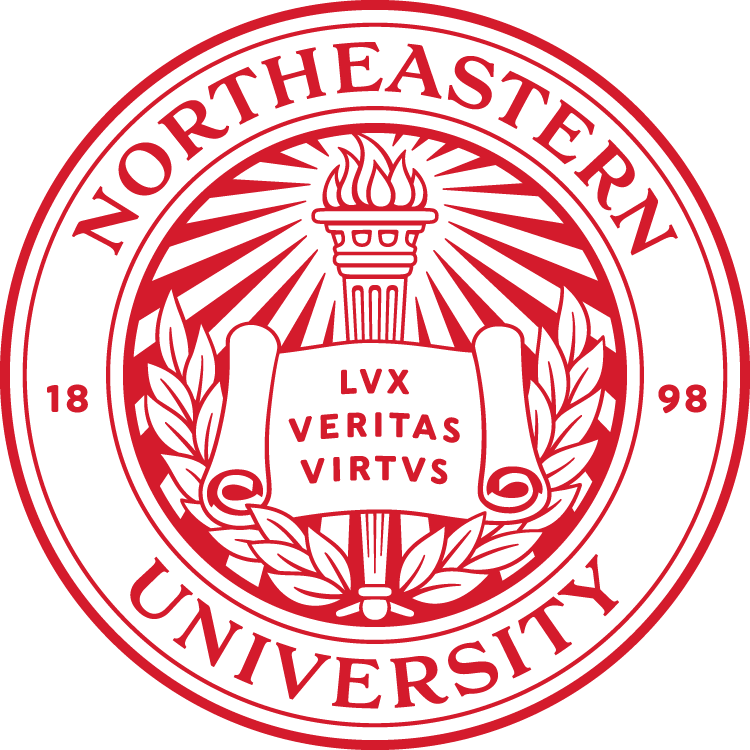}}\;
$^2$Northeastern University, Boston, MA, USA}

\authorrunning{Y. Pan et al.}

\maketitle

\vspace{-0.4cm}
\begin{center}
\small $^\star$Equal contribution
\end{center}


\vspace{-0.5cm}
\begin{abstract}
True video intelligence demands more than recognizing what is visible: it
requires reasoning about \emph{why} events unfold, predicting \emph{what would
change} under different conditions, and deciding \emph{what to do next}. We
refer to this full progression---from perception through causal reasoning and
simulation to strategic planning---as Strategic Video Intelligence (SVI). No
existing benchmark evaluates this capability stack: in-the-wild videos lack
verifiable ground truth for causal and strategic questions, while synthetic
environments sacrifice the complexity of real multi-agent systems. To bridge
this gap, we introduce SVI-Bench, a large-scale benchmark that leverages team
sports as a \emph{dynamic microworld}, a domain that uniquely combines the
complexity of real-world multi-agent interaction with the verifiability of
explicit rules and definitive outcomes. SVI-Bench comprises ${\sim}$\textbf{35K
hours} of broadcast video, ${\sim}$\textbf{15M} annotated actions,
${\sim}$\textbf{15K hours} of expert commentary, ${\sim}$\textbf{23K} game
reports, and ${\sim}$\textbf{103K} structured statistical records across
basketball, soccer, and hockey, all constructed via a data engine that
transforms raw game data into a dense, cross-referenced corpus. We organize
evaluation into \textbf{9 tasks} spanning a progressive four-pillar hierarchy:
\emph{Dynamic Scene Understanding}, \emph{Causal Reasoning}, \emph{Strategic
Simulation}, and \emph{Agentic Synthesis}. Evaluating strong multimodal and
agentic baselines, we find a \emph{capability cliff}: models perform competently on perceptual tasks (achieving ${\sim}$74\%
on fine-grained action QA) but degrade sharply at higher levels of the stack. Agentic tasks prove hardest of
all: the strongest model achieves only 5\% accuracy when required to
autonomously gather and integrate evidence across a corpus of 1.8M
clips. We release the full benchmark to catalyze progress toward
AI systems capable of strategic intelligence in complex, dynamic multi-agent
environments.
\end{abstract}
\vspace{-0.6cm}

\begin{center}
\small
\begin{tabular}{@{}r@{\hspace{1.5em}}l@{}}
\faGithub\;\textbf{Code} & \href{https://github.com/texaser/svi-bench}{\nolinkurl{github.com/texaser/svi-bench}} \\[2pt]
\faDatabase\;\textbf{Data} & \href{https://huggingface.co/mvp-group/svi-bench}{\nolinkurl{huggingface.co/mvp-group/svi-bench}} \\[2pt]
\faGlobe\;\textbf{Website} & \href{https://svi-bench.github.io}{\nolinkurl{svi-bench.github.io}} \\[2pt]
\faFile*\;\textbf{Extended Paper} & \href{https://svi-bench.github.io/svi_bench_extended.pdf}{\nolinkurl{svi-bench.github.io/svi_bench_extended.pdf}} \\[2pt]
\end{tabular}
\end{center}

\clearpage

\section{Introduction}
\label{sec:intro}
\vspace{-0.3cm}

In the final seconds of Game 6 of the 1998 NBA Finals, Michael Jordan receives the ball trailing by one.
Jordan perceives the defensive
formation shifting around him, infers that an aggressive first step will force
his defender to overextend, simulates how that reaction will open a path that
did not exist before, and selects the optimal action: a jump shot
that wins the championship. Jordan does not merely react to what he sees; he
reasons about its causes, anticipates its consequences, and synthesizes it all
into strategic action.

\begin{figure*}[t]
  \centering
\includegraphics[width=0.89\textwidth]{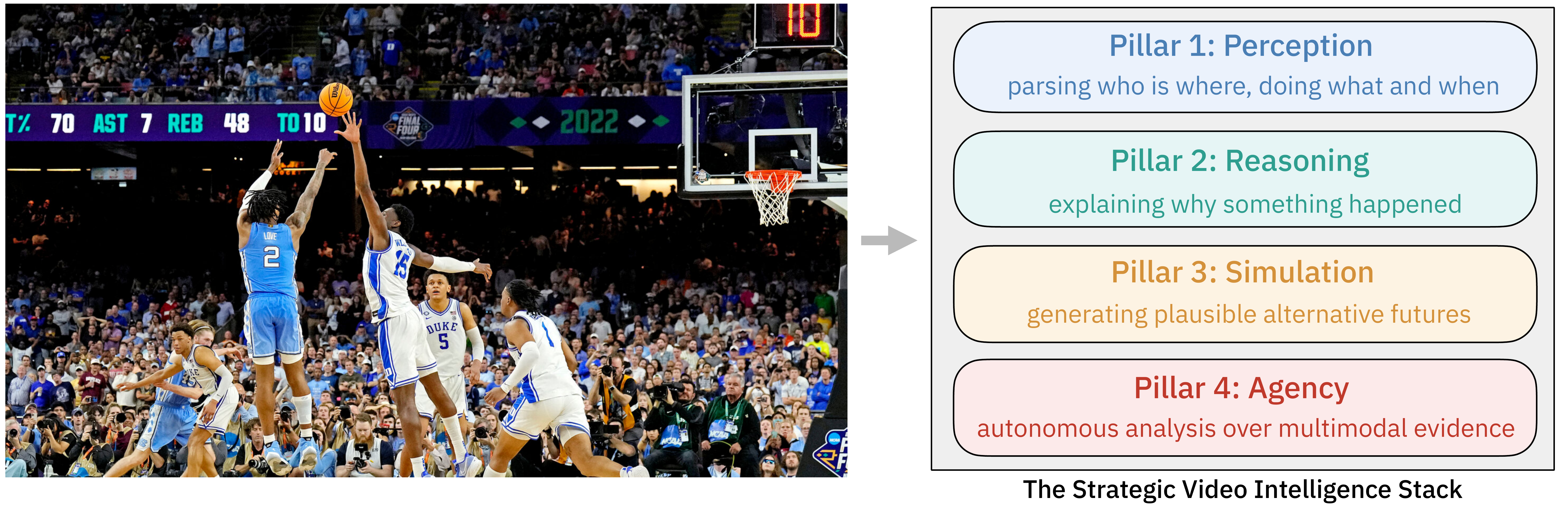}
\vspace*{-0.2cm}

  \caption{\textbf{Overview of \name{},} illustrated through a single play from the 2022 NCAA Final Four. \name{} is the first large-scale video benchmark evaluating the full SVI stack: Perception (describing what happens), Reasoning (explaining why), Simulation (generating plausible alternatives), and Agency (autonomous analysis).\vspace{-0.5cm}}

  \label{fig:teaser}
\end{figure*}

This kind of intelligence---moving from \emph{seeing} to
\emph{reasoning} to \emph{deciding}---remains out of reach for current AI systems.
A state-of-the-art video-language model, given the same footage, can describe
the scene: \emph{a player receives the ball, drives to the basket, releases a shot}. But it cannot explain \emph{why} the defense collapsed, predict \emph{what would have happened}
had the guard driven left instead of right, or recommend the \emph{optimal response} given the defensive
configuration. This gap extends well beyond sports: surgical
teams, first responders, autonomous vehicles, and military units all require
the same ability to reason about \emph{why} events unfold, simulate
\emph{what-if} alternatives, and decide \emph{what to
do next}.

We argue that these abilities are not independent skills but facets of an integrated capability that we call \textbf{Strategic Video Intelligence
(SVI)}: a progressive cognitive stack spanning \emph{perception} (parsing who is where and doing what), \emph{causal reasoning} (explaining why actions lead to outcomes),
\emph{simulation} (generating futures and
goal-directed strategies), and \emph{agentic synthesis} (autonomously
integrating multimodal evidence) (Figure~\ref{fig:teaser}).


Progress on SVI has been limited by the absence of
suitable benchmarks. Existing video
benchmarks~\cite{fu2025video,mangalam2023egoschema,zhou2025mlvu,wu2024longvideobench}
cover perception and temporal reasoning, but none evaluates
the full stack from perception to agency (Table~\ref{tab:benchmark_comparison}).
Synthetic datasets~\cite{yi2020clevrer,bakhtin2019phyre} provide verifiable ground truth for causal questions but without real-world complexity. Real-world benchmarks~\cite{fu2025video,mangalam2023egoschema}, by contrast, offer visual richness but no verifiable ground truth for causal or strategic reasoning.

Team sports offer a \emph{dynamic microworld} that bridges this gap.
Sports feature complex multi-agent dynamics (10 to 22 players executing
coordinated decisions under adversarial pressure) while offering properties
that make strategic reasoning measurable. First, \emph{long-horizon causality}:
early tactical setups (a screen, a formation shift) produce delayed outcomes
(a scoring opportunity, a turnover) seconds or minutes later, requiring models
to trace causal chains across extended temporal windows. Second,
\emph{unambiguous success signals}: scores, turnovers, and wins
provide clear outcome labels for causal and strategic questions. 


\begin{wrapfigure}[15]{r}{0.45\textwidth}
\begin{center}
\vspace{-1.25cm}
  \includegraphics[width=0.45\textwidth]{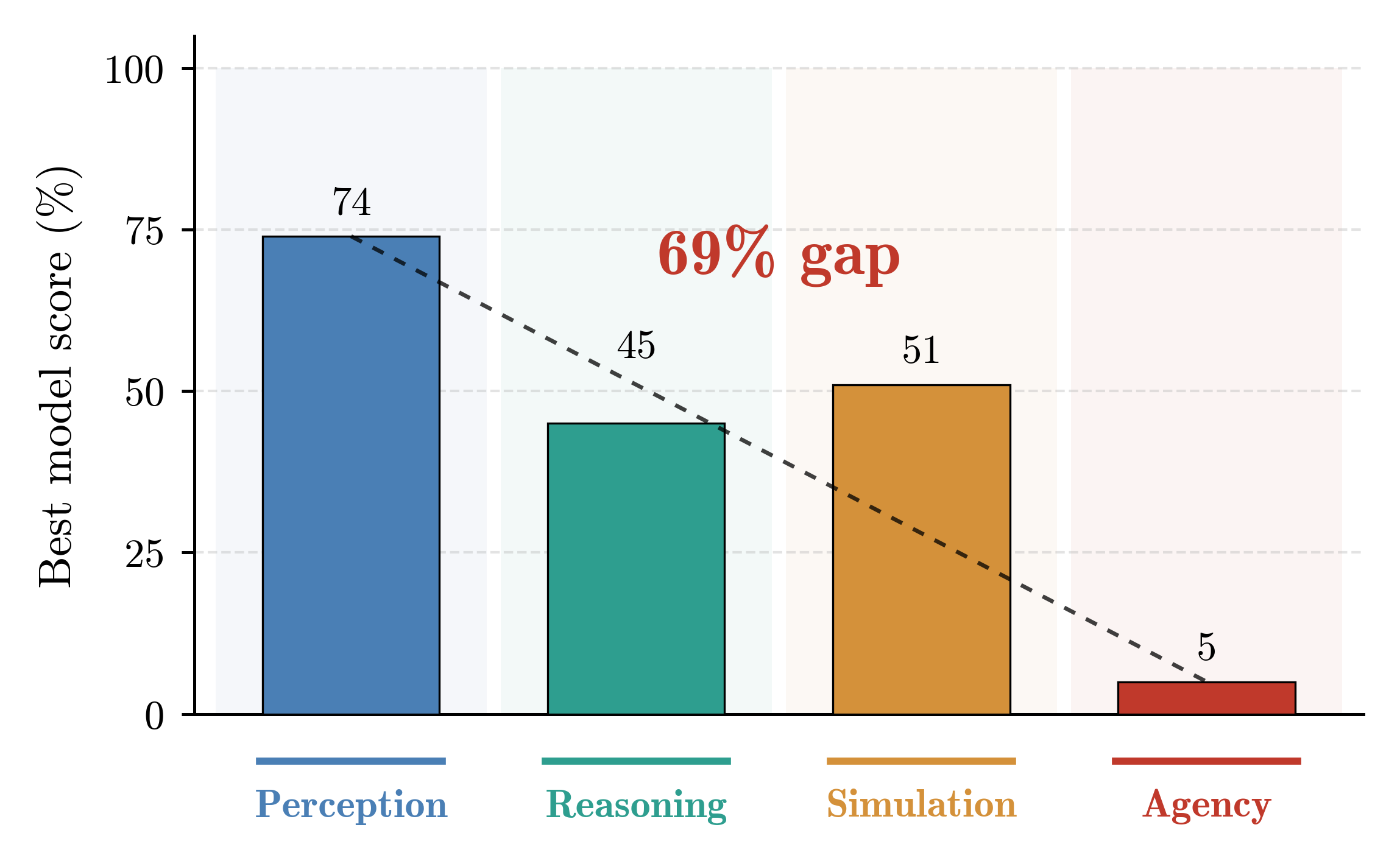}
\end{center}
\vspace{-0.7cm}
\caption{\textbf{The performance cliff.} Best task score per pillar, normalized to a 0--100\% range. From the strongest Perception result (74\%) to Agentic Synthesis (5\%), performance drops by 69 points, with Reasoning and Simulation falling in between.\vspace{-0.5cm}}
  \label{fig:capability_cliff}
\end{wrapfigure}

With this motivation, we introduce \name{}, the first large-scale benchmark
designed to evaluate the full SVI stack, from perception through reasoning and
simulation to agency, in real-world multi-agent video. \name{} consists of
${\sim}$35K hours of broadcast video, ${\sim}$15M annotated actions,
${\sim}$15K hours of expert commentary, ${\sim}$23K game reports, and
${\sim}$103K statistical records across basketball, soccer, and hockey (Table~\ref{tab:benchmark_comparison}). These five sources are integrated via a data
engine that performs temporal alignment, cross-modal entity resolution, and
LLM-powered instance generation with automated verification (\S\ref{sec:dataset}). We organize evaluation into 9 tasks across a progressive four-pillar hierarchy (\S\ref{sec:tasks}): (1)~\emph{Dynamic Scene Understanding}---parsing multi-agent scenes into structured spatiotemporal representations; (2)~\emph{Causal Reasoning}---explaining why events unfold and predicting outcomes; (3)~\emph{Strategic Simulation}---generating counterfactual futures and goal-directed strategies; and (4)~\emph{Agentic Synthesis}---autonomously gathering and integrating multimodal evidence to produce expert-level analysis. Evaluating strong multimodal and agentic baselines, we find that models perform competently on perception but decline sharply on reasoning and agentic tasks (Figure~\ref{fig:capability_cliff}). Our contributions are:

\vspace{-0.25cm}
\begin{enumerate}[leftmargin=*,itemsep=1pt]
\item \textbf{The Strategic Video Intelligence framework}, formalizing video understanding as a progressive perception-to-agency stack.
\item \textbf{A data engine} that aligns five modalities via temporal alignment, cross-modal entity resolution, LLM-assisted generation, and quality control.
\item \textbf{\name{}}, a large-scale benchmark spanning the full perception-to-agency stack, with 9 tasks across the four pillars and training splits for 7.
\item \textbf{Reference methods and analysis} for every task, localizing where and why performance degrades.
\end{enumerate}

\begin{table*}[t]
\centering
\caption{\textbf{Comparison with existing video benchmarks.} \name{} is the first to combine large-scale, real-world multi-agent video with cross-referenced multimodal data and evaluation spanning perception through strategic agency.}
\vspace{-0.3cm}
\label{tab:benchmark_comparison}
\normalsize
\setlength{\tabcolsep}{3pt}
\renewcommand{\arraystretch}{1.1}
\begin{adjustbox}{max width=\textwidth}
\begin{tabular}{@{}cl r r r r r r cccc@{}}
\toprule
& & & & & \multicolumn{3}{c}{\textbf{Modalities}} & \multicolumn{4}{c}{\textbf{Evaluation Tasks}} \\
\cmidrule(lr){6-8} \cmidrule(lr){9-12}
&\textbf{Benchmark}
 & \makecell[c]{\textbf{Video}\\\textbf{Hours}}
 & \makecell[c]{\textbf{Duration}\\\textbf{Range}}
 & \makecell[c]{\textbf{Annotated}\\\textbf{Actions}}
 & \makecell[c]{\textbf{Expert}\\\textbf{Commentary}}
 & \makecell[c]{\textbf{Long-Form}\\\textbf{Reports}}
 & \makecell[c]{\textbf{Structured}\\\textbf{Metadata}}
 & \rotatebox{60}{\textbf{Perception}}
 & \rotatebox{60}{\textbf{Reasoning}}
 & \rotatebox{60}{\textbf{Simulation}}
 & \rotatebox{60}{\textbf{Agency}}
 \\
\midrule
\multirow{5}{*}{\rotatebox{90}{\textcolor{sectiongray}{\textit{\small General}}}}
& Kinetics-700  & 1.9K      & 10s         & 650K  & \xmark & \xmark & \xmark & \cmark & \xmark & \xmark & \xmark \\
& ActivityNet   & 849       & 5--10 min   & 30K   & \xmark & \xmark & \xmark & \cmark & \xmark & \xmark & \xmark \\
& Video-MME     & 254       & 11s--1h    & --    & \xmark & \xmark & \xmark & \cmark & \pmark & \xmark & \xmark \\
& Ego-Exo4D     & 1.4K      & 1--42 min   & 432K    & 6K hrs & \xmark & \xmark & \cmark & \xmark & \xmark & \xmark \\
& Ego4D-HCap    & 3.7K      & 5s--2h      & 3.8M & \xmark & 8K & \xmark & \cmark & \xmark & \xmark & \xmark \\
\midrule
\multirow{3}{*}{\rotatebox{90}{\textcolor{sectiongray}{\textit{\small Reasoning}}}}
& Causal-VidQA  & 28        & 10s         & --    & \xmark & \xmark & \xmark & \cmark & \pmark & \xmark & \xmark \\
& MINERVA       & $\sim$150 & 2--100 min   & --    & \xmark & \xmark & \xmark & \cmark & \cmark & \xmark & \pmark \\
& Video-Holmes  & $\sim$14  & 1--5 min    & --    & \xmark & \xmark & \xmark & \cmark & \cmark & \xmark & \pmark \\
\midrule
\multirow{2}{*}{\rotatebox{90}{\textcolor{sectiongray}{\textit{\small Synth.}}}}
& CLEVRER       & 5         & 5s          & --    & \xmark & \xmark & \xmark & \cmark & \cmark & \pmark & \xmark \\
& PHYRE         & --        & 5s          & --    & \xmark & \xmark & \xmark & \pmark & \cmark & \pmark & \xmark \\
\midrule
\multirow{3}{*}{\rotatebox{90}{\textcolor{sectiongray}{\textit{\small Sports}}}}
& SoccerNet     & 500       & 90 min      & 300K  & \xmark & \xmark & 500 & \cmark & \xmark & \xmark & \xmark \\
& SportsMOT     & 14        & 14.4--33.8s & --    & \xmark & \xmark & \xmark & \cmark & \xmark & \xmark & \xmark \\
& BASKET        & 4.5K      & 8--10 min   & --    & \xmark & \xmark & \xmark & \cmark & \xmark & \xmark & \xmark \\
\midrule
\rowcolor{oursrow}
& \textbf{\name{} (Ours)} & \textbf{35K} & \textbf{10s--2.5h} & \textbf{15M} & \textbf{15K hrs} & \textbf{23K} & \textbf{103K} & \cmark & \cmark & \cmark & \cmark \\
\bottomrule
\end{tabular}
\end{adjustbox}
\vspace{-0.5cm}
\end{table*}


\vspace{-0.1cm}
\begin{tcolorbox}[
  enhanced, breakable, sharp corners,
  colback=gray!4, colframe=gray!55,
  boxrule=0.4pt, left=8pt, right=8pt, top=2pt, bottom=2pt,
  before skip=0pt, after skip=0pt,
  fontupper=\small
]
\textbf{Extended version.} This is the compact version of the paper. An \href{https://svi-bench.github.io/svi_bench_extended.pdf}{extended version} provides full dataset construction details, per-task statistics, task-specific prompts, additional analysis, and complete results for every task.
\end{tcolorbox}

\vspace{-0.25cm}
\section{Related Work}
\label{sec:related}
\vspace{-0.15cm}

\begin{description}[style=unboxed,leftmargin=0pt,itemsep=2pt,parsep=0pt,topsep=3pt]


\item[\textbf{Video Benchmarks.}]
Early benchmarks targeted action recognition~\cite{soomro2012ucf101,kay2017kinetics,carreira2019short}
and temporal localization~\cite{caba2015activitynet,idrees2017thumos}. Video QA benchmarks~\cite{xu2017video,yu2019activitynet,xiao2021next,mangalam2023egoschema}
added language-grounded reasoning over short clips with predominantly perceptual questions. Long-video benchmarks~\cite{fu2025video,zhou2025mlvu,wu2024longvideobench} extend temporal scope but lack verifiable causal ground truth. Egocentric planning~\cite{chen2026egoplan} and temporal reasoning~\cite{liu2024tempcompass,li2026timeblind} benchmarks target single-agent or low-complexity settings.

\item[\textbf{Causal and Counterfactual Reasoning in Video.}]
Synthetic environments~\cite{yi2020clevrer,bakhtin2019phyre,baradel2020cophy,bear2021physion} provide verifiable ground truth for causal and physical reasoning but involve single objects in simplified worlds without multi-agent behavioral complexity.
Real-video causal QA~\cite{li2022causalvidqa,mecd_v1} attempts causal reasoning but relies on subjective annotations without objectively verifiable outcomes.

\item[\textbf{Sports Video Analysis.}]
Prior work targets action detection~\cite{giancola2018soccernet,deliege2021soccernet,cioppa2022soccernet}, tracking~\cite{cui2023sportsmot}, trajectory prediction~\cite{alcorn2021baller2vec,felsen2018where}, and skill analysis~\cite{xu2024finesports,pan2025basket}. SPORTU~\cite{xia2024sportu} and SportR~\cite{xia2025sportr} add rule comprehension but omit simulation or agentic reasoning. NBA tracking datasets~\cite{cervone2016multiresolution} lack video or language. 

\item[\textbf{Multimodal LLMs and Agentic AI.}]
Recent multimodal LLMs~\cite{openai2024gpt4ocard,geminiteam2025geminifamilyhighlycapable,liu2024llava,lin2023videollava,zhang2023videollama}
demonstrate strong video description but exhibit brittle temporal reasoning~\cite{fu2025video,mangalam2023egoschema}. Agentic AI systems~\cite{yao2023react,schick2023toolformer,wang2024voyager} combine LLMs with tool use, and recent work extends this to video~\cite{timesearch-r,videothinker,yang2025longvt,thinkingwithvideos,rasheed2025video}. \name{}'s Pillar~4 introduces the first agentic video evaluation at corpus scale.


\item[\textbf{Strategic Game AI and World Models.}]
Superhuman game AI~\cite{silver2017mastering,schrittwieser2020muzero,vinyals2019grandmaster} operates in fully observable, discrete-state environments. \name{} targets continuous, partially observable, real multi-agent video. World models span model-based RL~\cite{ha2018world,hafner2020dreamerv2,hafner2023dreamerv3}, diffusion~\cite{alonso2024diamond}, transformer~\cite{robine2023transformerwm,micheli2023iris}, and foundation~\cite{ho2022video,bruce2024genie,agarwal2025cosmos} approaches, but none target strategic reasoning in multi-agent video.



\end{description}

\vspace{-0.4cm}
\section{Data Engine}
\label{sec:dataset}
\vspace{-0.2cm}


A core contribution of \name{} is a data engine that transforms raw game data into a dense, cross-referenced corpus for evaluating the full SVI stack. It is built on two principles: (i)~primary evidence is human- or league-derived (play-by-play logs, official statistics, journalist reports, broadcast commentary), and (ii)~LLMs \emph{scale} task-instance generation from these grounded sources, with manual verification on a representative subset of every task. This yields supervision at a scale and density unmatched by prior sports video resources.

\vspace{-0.3cm}
\subsection{Data Sources and Scale}
\label{sec:sources}

\begin{figure*}[t]
  \centering
\includegraphics[width=0.92\textwidth]{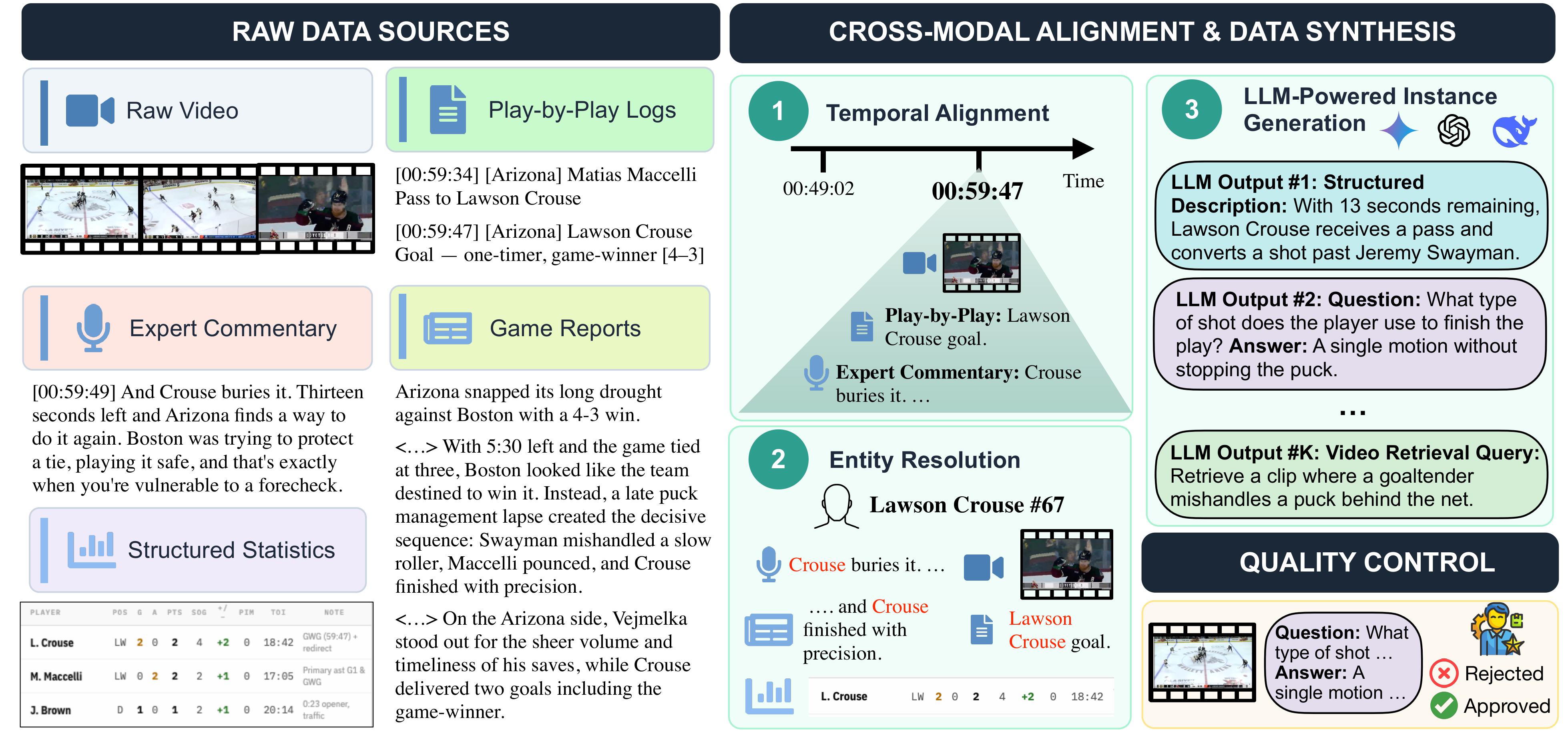}
\vspace*{-0.2cm}

\caption{\textbf{The \name{} data engine.} Five raw sources are transformed into a cross-referenced corpus via (1)~temporal alignment, (2)~cross-modal entity resolution, (3)~LLM-based instance generation, and (4)~automatic and human quality control.\vspace{-0.5cm}}
  \label{fig:data_engine}
\end{figure*}

\name{} spans three team sports selected for their
complementary multi-agent properties in team size, pacing, spatial scale,
camera dynamics, and strategic structure: basketball (10 players, compact
court, frequent transitions), soccer (22 players, large pitch, continuous
fluid dynamics), and hockey (rapid line changes, fast-panning camera). The
corpus comprises five synergistic modalities (Figure~\ref{fig:data_engine},
left), all temporally aligned and cross-referenced through shared game and
player identifiers:
\textbf{broadcast video} (${\sim}$35K hours spanning multiple seasons);
\textbf{play-by-play logs} (${\sim}$15M timestamped event records from official league data feeds, including shots, passes, fouls, and substitutions with player identities and spatial coordinates);
\textbf{expert commentary} (${\sim}$15K hours of broadcast commentary and analyst narration collected via ASR);
\textbf{game reports} (${\sim}$23K post-game journalist recaps and editorial analyses); and
\textbf{box-score statistics} (${\sim}$103K records of player/team performance metrics and standings).

\suppressfloats[t]
\begin{table}[t]
\centering
\caption{\textbf{Summary of \name{} benchmark tasks.} Nine tasks
across four cognitive pillars covering basketball (B), hockey (H), and soccer (S).
\# is total task instances. Train indicates whether a training split
is provided.}
\vspace{-0.2cm}
\label{tab:task_summary}
\footnotesize
\setlength{\tabcolsep}{3pt}
\renewcommand{\arraystretch}{1.0}
\begin{adjustbox}{max width=\columnwidth}
\begin{tabular}{@{}c l l c c r c l l@{}}
\toprule
\textbf{ID} & \textbf{Task} & \textbf{Pillar} & \textbf{Context} & \textbf{Sports} & \textbf{\#} & \textbf{Train} & \textbf{Format} & \textbf{Primary Metric} \\
\midrule
T1 & Structured Play Description       & \textcolor{pillar1}{\textbf{Perception}}  & 10s         & B,H,S & 1.5M & \cmark & Open       & Avg. Score  \\
T2 & Fine-Grained Action QA            & \textcolor{pillar1}{\textbf{Perception}}  & 10s         & B,H,S & 1.5M & \cmark & MCQ        & Accuracy    \\
T3 & Compositional Video Retrieval     & \textcolor{pillar1}{\textbf{Perception}}  & 10s         & B,H,S & 306K & \cmark & Retrieval  & R@1         \\
\addlinespace[2pt]
T4 & Strategic Reasoning QA            & \textcolor{pillar2}{\textbf{Reasoning}}   & 55--150 min & B,H,S     & 1K   & \xmark & Open       & Avg. Score  \\
T5 & Outcome Forecasting               & \textcolor{pillar2}{\textbf{Reasoning}}   & 3--15 min   & B,H,S & 114K & \cmark & MCQ        & Accuracy    \\
T6 & Long-form Narrative Synthesis     & \textcolor{pillar2}{\textbf{Reasoning}}   & 55--150 min & B,H,S & 19K  & \cmark & Open       & Saliency    \\
\addlinespace[2pt]
T7 & Motion-Conditioned Generation     & \textcolor{pillar3}{\textbf{Simulation}}  & 5--10s         & B,S   & 290K & \cmark & Generation & Video mIoU  \\
T8 & Goal-Conditioned Action Gen.      & \textcolor{pillar3}{\textbf{Simulation}}  & 5--10s         & B     & 74K  & \cmark & Generation & Goal Acc.   \\
\addlinespace[2pt]
T9 & Cross-Corpus Agentic Reasoning    & \textcolor{pillar4}{\textbf{Agency}}      & Multi Source& B,H,S     & 1K   & \xmark & Open       & Accuracy    \\
\bottomrule
\end{tabular}
\end{adjustbox}
\vspace{-0.4cm}
\end{table}

\vspace{-0.3cm}
\subsection{Data Construction Pipeline}
\label{sec:engine}

Our data engine (Figure~\ref{fig:data_engine}) transforms these five raw sources into a cross-referenced corpus through four stages.
\textbf{(1)~Temporal alignment:} Play-by-play logs provide the primary temporal reference via game-clock timestamps; commentary transcripts and game reports are aligned using timestamp matching and textual cues, producing temporally grounded segments linking every video clip to its corresponding events, commentary, and statistical context.
\textbf{(2)~Cross-modal entity resolution:} References to the same player, team, or event are linked across modalities and organized into identity graphs capturing relationships (teammate, opponent) and attributes (position, statistics, role).
\textbf{(3)~LLM-powered instance generation:} Using the assembled multimodal context, LLMs synthesize instances guided by pillar-aware prompt templates, generating question--answer pairs, plausible distractors, difficulty-calibrated instances, and free-form annotations such as dense captions and narrative summaries.
\textbf{(4)~Quality control:} All instances pass through automatic checks against event logs, followed by human expert review by domain-knowledgeable annotators across a balanced subset spanning all sports, pillars, and difficulty levels.

\vspace{-0.3cm}
\section{The \name{} Evaluation Suite}
\label{sec:tasks}
\vspace{-0.1cm}

SVI-Bench comprises 9 tasks organized into a
four-pillar hierarchy (Table~\ref{tab:task_summary}), from perception through causal reasoning and simulation to
agency. Construction details, prompts, per-task statistics, and complete results are in the extended version. Below, we present each pillar and its tasks.

\vspace{-0.3cm}
\subsection{Pillar 1: Dynamic Scene Understanding (T1--T3)}
\label{sec:pillar1}
\vspace{-0.1cm}

\begin{figure*}[t]
  \centering
\includegraphics[width=0.89\textwidth]{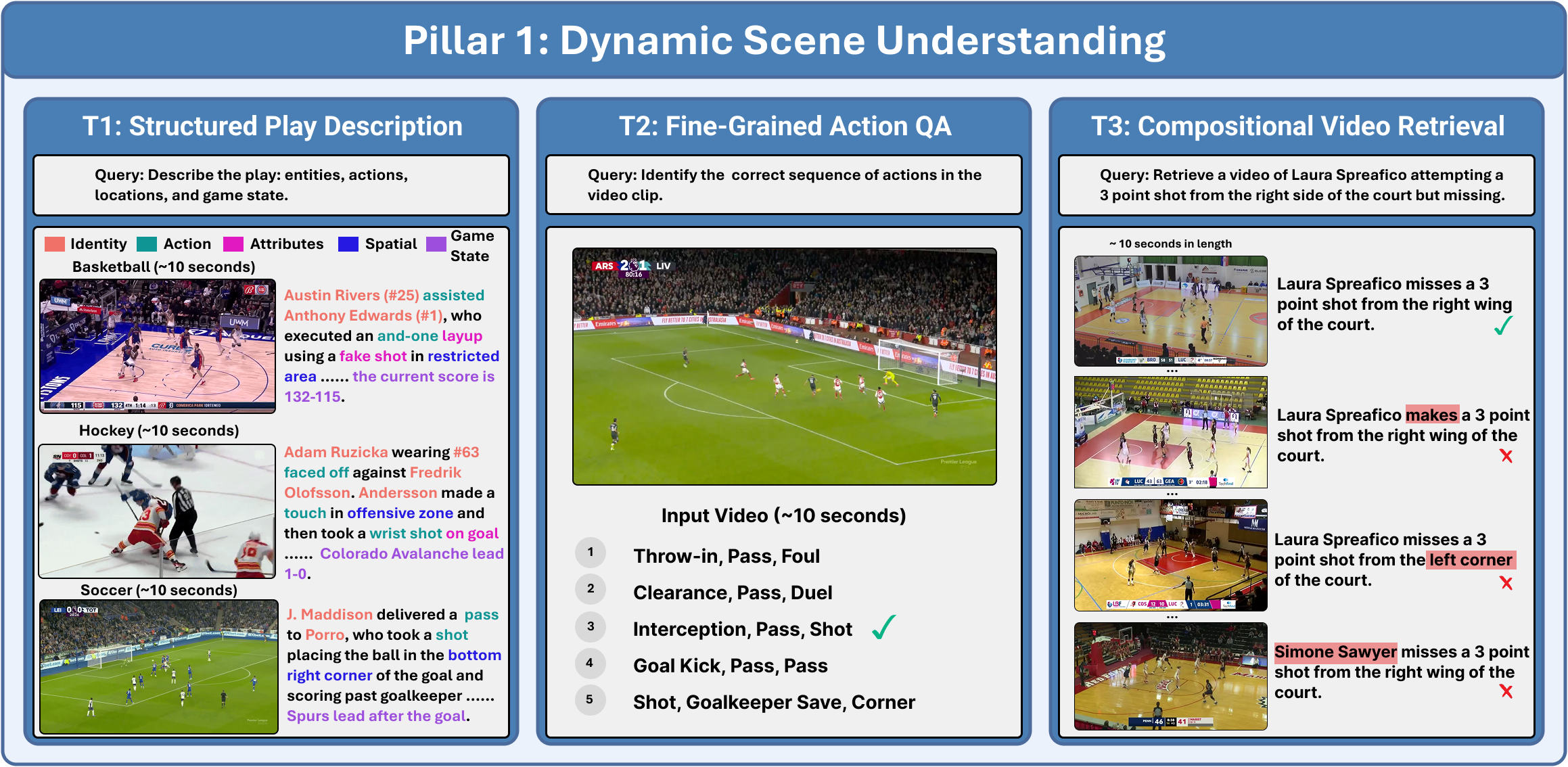}
\vspace{-0.1cm}

  \caption{\textbf{Overview of Pillar 1: Dynamic Scene Understanding}. This pillar evaluates foundational perceptual capabilities through three tasks: structured play description (T1), fine-grained action QA (T2), and compositional video retrieval (T3).\vspace{-0.6cm}}

  \label{fig:pillar1_examples}
\end{figure*}

Strategic reasoning begins with perception, parsing a dense, fast-moving multi-agent scene into spatiotemporal primitives: which agents are present, where they are, what they are doing, and how these compose into higher-order events. The three tasks in this pillar evaluate this foundation using short clips, establishing the perceptual floor upon which higher-level reasoning depends (Figure~\ref{fig:pillar1_examples}).

\suppressfloats[t]

\subsubsection{T1: Structured Play Description}

\mypara{Task formulation.}
Given a 10-second video, the model must generate a
dense, structured caption that describes the actions, player identities,
spatial positioning, and game context. Unlike standard captioning benchmarks~\cite{krishna2017densecaptioning,xu2016msrvttcaption,wang2019vatex,zhou2018youcook2} describing a single salient action per clip, T1 requires describing dynamic scenes with 10+ coordinated agents, parallel sub-actions, and game-state context.

\mypara{Data and construction.}
We extract 10-second segments centered on specific events, with captions synthesized from play-by-play annotations and refined via GPT-4o-mini for linguistic diversity and narrative coherence.

\mypara{Evaluation.}
We use an LLM-as-a-judge protocol with scoring rubrics, assigning
Likert scores from 0--5 along six axes: action, identity,
causality/outcome, spatial understanding, temporal understanding, and contextual
details. We report the
average score across all axes as the primary metric.  To verify judge reliability, three annotators independently scored 60 randomly sampled instances. The mean absolute difference between human and LLM-judge scores is 0.40 on the 0--5 scale, indicating agreement within normal inter-annotator variation.


\mypara{Baselines and findings.}
GPT-5.2 achieves
only 1.61/5.00 average score, and Gemini 3 Flash reaches 1.67. Fine-tuned LLaVA-Video-7B reaches 2.17
(+1.28 over zero-shot), demonstrating the value of domain-specific training.
Per-axis analysis reveals that models perform relatively well on spatial
understanding (2.74) and temporal understanding (2.72) but struggle with
identity recognition (1.11) and causality/outcome reasoning (1.82).

\subsubsection{T2: Fine-Grained Action QA}

\mypara{Task formulation.}
Given a 10-second clip, a question, and 5
candidate answers, the model must select the correct one. Unlike prior video QA benchmarks~\cite{xu2017video,yu2019activitynet,xiao2021next,mangalam2023egoschema} that feature single-agent scenarios with coarse-grained questions, T2 targets multi-agent interactions where correct answers depend on precise details (e.g., which player initiated a screen, the exact pass sequence, or spatial relationships). The task spans six categories (action recognition, temporal ordering, play analysis, spatial relationships, player identification, and OCR).

\mypara{Data and construction.}
We segment full-game footage into 10-second clips centered on individual plays
with dense play-by-play annotations, covering 31 question types across three sports organized into six categories.

\mypara{Evaluation.}
We report average accuracy across all sports and
question types.


\mypara{Baselines and findings.}
Gemini 3 Flash achieves 58.75\% accuracy while GPT-5.2 achieves 52.91\%. Fine-tuned LLaVA-Video-7B reaches
73.91\% (+36.9\% over zero-shot). Player identification questions are most challenging, while action recognition is easier. Sport-experienced humans reach 75.78\% overall (Section~\ref{sec:cross_task}).


\subsubsection{T3: Compositional Video Retrieval}

\mypara{Task formulation.}
Given a natural-language query describing a specific composition of visual
attributes (entity, dynamics, context, spatiotemporal structure) and a
candidate pool of one positive and 5{,}000 negative videos, the model must
rank the candidate videos by semantic similarity with the query, aiming to
place the ground-truth video at the top. Unlike standard video retrieval~\cite{xu2016msrvttcaption,anne2017didemo,krishna2017densecaptioning,rohrbach2015lsmdc} where visually diverse candidates allow coarse features to suffice, all T3 candidates depict the same sport and many share nearly identical visual elements, differing only in their specific composition of attributes.

\mypara{Data and construction.}
Queries are generated from ground-truth attributes of each video and refined
into natural language via LLM paraphrasing, with hard-negative mining to ensure challenging distractors.

\mypara{Evaluation.}
We report Recall@$K$ with R@1 as the primary metric.

\mypara{Baselines and findings.}
We fine-tune InternVideo2\cite{wang2024internvideo2} using a video-text contrastive loss.
The model achieves an aggregate R@1 of 3.0\%
and R@10 of 13.3\%, highlighting the difficulty of fine-grained retrieval at scale. When the
proportion of near-duplicate distractors increases,
R@100 drops by nearly half, confirming that distinguishing visually similar compositions remains a core challenge.

\begin{finding}{pillarblue}{Pillar 1 Takeaway (Dynamic Scene Understanding).}
Perception is the strongest pillar, yet significant gaps remain:
fine-grained action QA reaches 73.91\%, while structured
captioning and compositional retrieval reveal persistent weaknesses in
identity grounding and multi-attribute
composition.
\end{finding}

\vspace{-0.3cm}
\subsection{Pillar 2: Causal Reasoning (T4--T6)}
\label{sec:pillar2}
\vspace{-0.1cm}



\begin{figure*}[t]
  \centering
\includegraphics[width=0.89\textwidth]{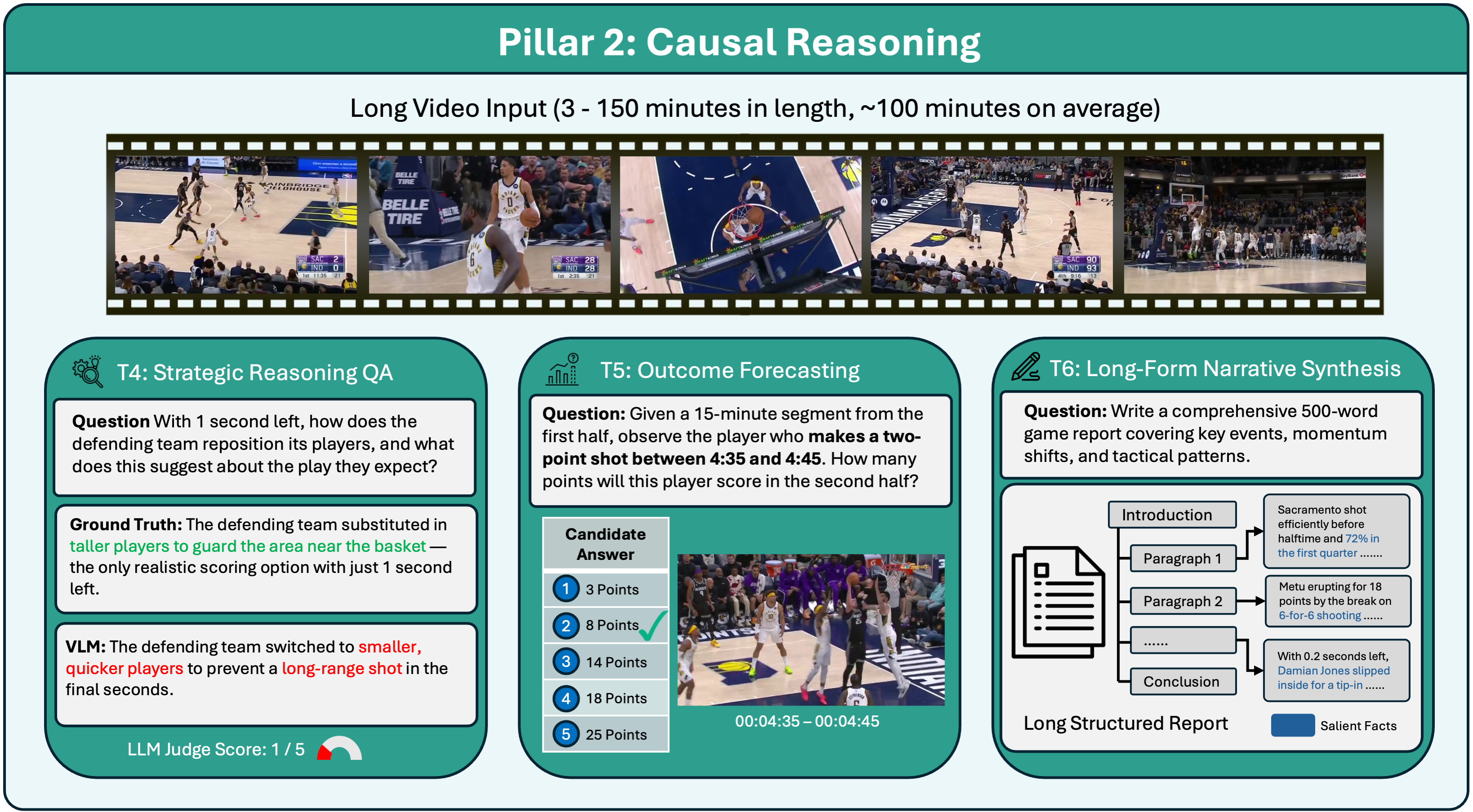}
\vspace{-0.2cm}
  \caption{\textbf{Overview of Pillar 2: Causal Reasoning.} This pillar evaluates the ability to reason about game-level context through three tasks: strategic reasoning QA (T4), outcome forecasting (T5), and long-form narrative synthesis (T6).\vspace{-0.8cm}}
  \label{fig:pillar2}
\end{figure*}

This pillar evaluates causal reasoning over long-form video spanning minutes to hours, testing whether models can explain the causes behind game events (T4), predict how a situation will unfold (T5), and synthesize extended developments into coherent narratives (T6) (Figure~\ref{fig:pillar2}).

\subsubsection{T4: Strategic Reasoning QA}

\mypara{Task formulation.}
Given a full-game video (${\sim}$55--150 min) and a question,
the model must produce a free-form response explaining strategic reasoning
behind game events. Unlike T2, which tests localized perception over short
clips, T4 requires reasoning over extended game portions: identifying
strategic errors, evaluating tactical execution, and interpreting
dynamics such as momentum shifts. Compared to long-form video QA benchmarks~\cite{tapaswi2016movieqa,fu2025video,zhou2025mlvu} whose questions are predominantly perceptual, T4 targets strategic causal reasoning where evidence may be spread across minutes of footage interleaved with irrelevant events.

\mypara{Data and construction.}
We curate 1{,}000 questions from expert commentaries and
game reports across all three sports, totaling 825 unique games.
A multi-stage pipeline generates open-ended question--answer pairs
followed by bias-mitigation filtering and human validation of factual validity and question quality.

\mypara{Evaluation.}
We use an LLM-as-a-judge protocol that scores each
response on a 0--5 scale, assessing factual consistency
with the reference answer and reasoning coherence. To validate judge reliability, a human annotator scored all model responses for 25 evaluation instances. The mean score difference between human and LLM judge is 0.12 on the 0--5 scale, confirming strong alignment. 


\mypara{Baselines and findings.}
We evaluate proprietary models (GPT-5.2, Gemini~3.1 Pro) and
open-source models (Qwen3-VL-32B, Molmo~2-8B). All models score near 2/5 on
average. Gemini~3.1 Pro is strongest overall (2.17), driven primarily by
soccer (2.49); GPT-5.2 (2.06) leads on basketball (1.99). Qwen3-VL-32B and
Molmo~2-8B reach 2.01 and 1.82 respectively. These low scores reflect the difficulty of T4, which requires reasoning about causal relationships and strategic intent, not just recognizing individual events.



\vspace{-0.2cm}
\subsubsection{T5: Outcome Forecasting}
\vspace{-0.1cm}

\mypara{Task formulation.}
Given a video segment capturing a sequence of play (3--15 minutes) and a
question about a future event, the model must predict the outcome by selecting
the correct answer from a candidate set. The target event occurs beyond the input window, requiring the model to infer the most
probable course of game development. Unlike trajectory forecasting~\cite{gupta2018social,liang2020garden,mangalam2020not,salzmann2020trajectron} that predicts short-horizon spatial positions, T5 targets semantically rich outcomes (who will score, which strategy will be used, how a game state will evolve) requiring understanding of complex causal mechanisms rather than trend extrapolation. Questions span \emph{performance forecasting} (predicting player or team
statistical accomplishments), \emph{game state evolution} (anticipating scores and possessions), and \emph{strategic intention} (identifying
the most probable tactical shifts).

\mypara{Data and construction.}
We curate 114K multiple-choice questions spanning 15 question types across
three sports, using game videos and dense play-by-play event annotations with
video segments ranging from 3 to 15 minutes.

\mypara{Evaluation.}
We use top-1 accuracy and additionally report calibration error to measure alignment between predicted confidence and empirical
correctness.

\mypara{Baselines and findings.}
We benchmark open-source (Qwen3-VL-8B~\cite{bai2025qwen3vltechnicalreport}, Molmo~2-8B~\cite{clark2026molmo2},
BIMBA~\cite{islam2025bimba}) and proprietary models (GPT-5.2, Gemini~3.0 Pro). No model exceeds
43.18\% under zero-shot evaluation. Fine-tuning Qwen3-VL-8B improves accuracy to 44.82\% (+7.9\% over zero-shot). Even frontier models like GPT-5.2 are poorly calibrated, with confidence exceeding accuracy by 28 points. 




\vspace{-0.1cm}
\subsubsection{T6: Long-Form Narrative Synthesis}

\mypara{Task formulation.}
Given a full game video (${\sim}$55--150 min) and a writing
prompt, the model must synthesize a narrative report
(${\sim}$500 words) covering key events, standout performances, and
strategic developments. Unlike video summarization methods~\cite{zhong2021qmsum,chen2022summscreen,papalampidi2020screenplay} that rely on dialogue or narration, T6 requires narratives grounded entirely in visual evidence from hours of multi-agent interaction, demanding saliency and factual precision across extreme temporal scales.

\mypara{Data and construction.}
We define 10 report templates per sport
(e.g., game narrative, player impact, team strategy evolution), five targeting single-game analysis and five requiring synthesis across multiple games. Reference reports are generated by an LLM from play-by-play logs, box scores, and journalist reports, ensuring consistent structure and verifiable factual content.

\mypara{Evaluation.}
We evaluate along three metrics using an LLM-as-a-judge (Qwen3-235B
Thinking \cite{qwen3technicalreport}): \emph{factual accuracy} via atomic fact decomposition~\cite{min2023factscore}; \emph{saliency}, measuring coverage of key events and performances as identified by a state-of-the-art LLM given oracle game information (play-by-play logs, box scores, and original journalist reports); and
\emph{writing quality}, rated on a 1--5 scale for coherence, topic adherence,
and length compliance.

\mypara{Baselines and findings.}
We benchmark Qwen3-VL-8B, GPT-5, and Gemini~3.1 Pro. Models achieve
relatively high factual accuracy at 73.01\%,
but struggle with saliency (7.33\%). Writing quality is the least problematic dimension, with most models scoring above 4.5/5.



\begin{finding}{pillarteal}{Pillar 2 Takeaway (Causal Reasoning).}
Models degrade sharply on reasoning tasks. T4 strategic reasoning scores 2/5 and T5 forecasting stays below 45\%. On T6 models reach 73.0\% factual accuracy but only 7.33\% saliency, describing what happened while failing to cover salient events. Across all three tasks, causal and strategic reasoning emerges as the main bottleneck.
\end{finding}

\vspace{-0.2cm}
\subsection{Pillar 3: Strategic Simulation (T7--T8)}
\label{sec:pillar3}

\begin{figure*}[t]
  \centering
\includegraphics[width=0.89\textwidth]{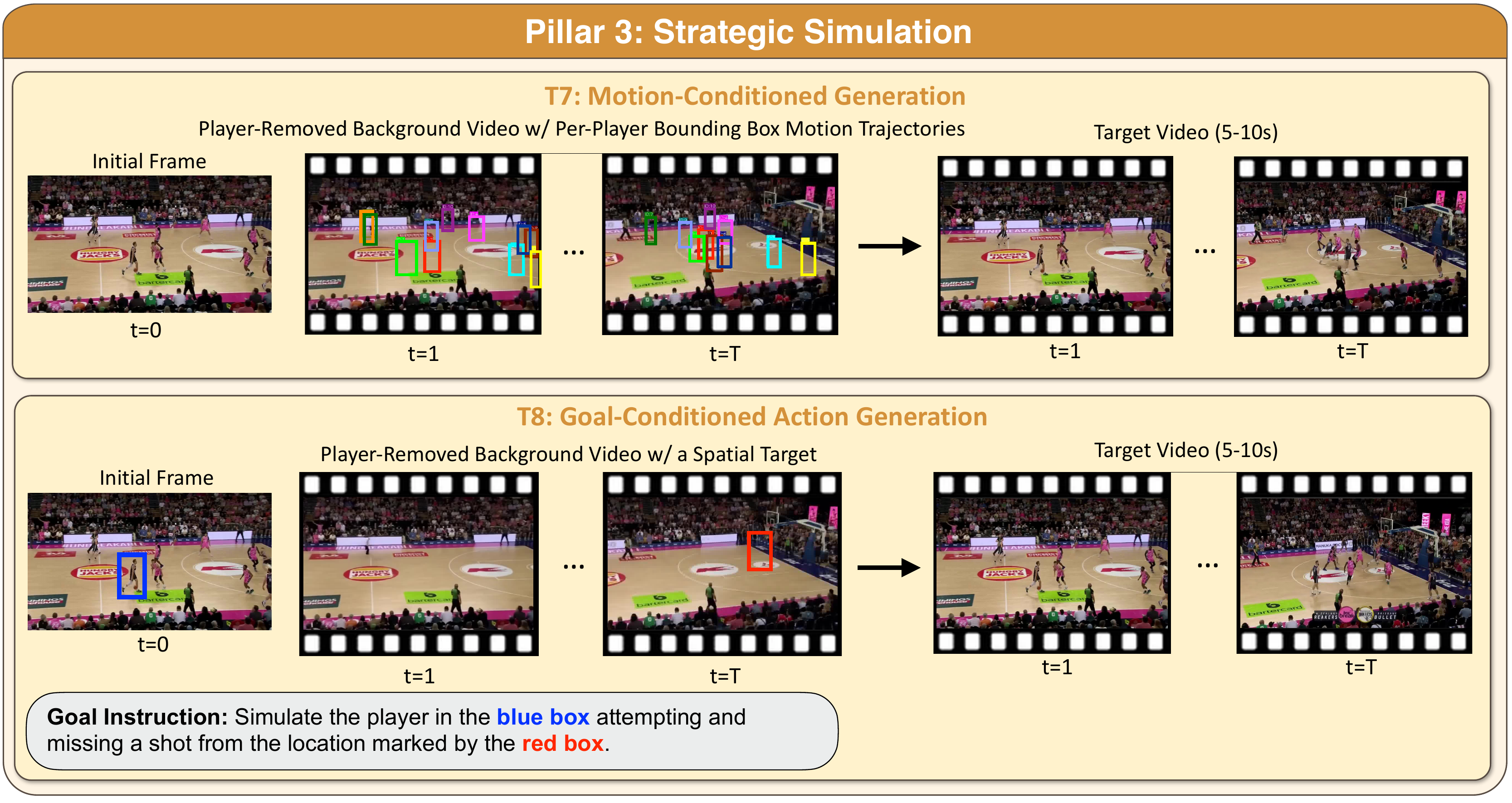}
\vspace*{-0.2cm}

  \caption{\textbf{Overview of Pillar 3: Strategic Simulation.} This pillar tests the ability to simulate alternative futures through two video generation tasks: motion-conditioned generation (T7), where players follow prescribed trajectories, and goal-conditioned action generation (T8), where the model plans actions toward a specified goal.\vspace{-0.3cm}}
  \label{fig:pillar3}
\end{figure*}

Understanding why something happened is distinct from reasoning about what \emph{could} happen. This pillar evaluates whether models can simulate alternative futures through video generation. Given a short clip (5--10s), both tasks require generating a realistic video of how the scene evolves if players follow specified trajectories (T7) or execute an action toward a goal (T8) (Figure~\ref{fig:pillar3}).

\subsubsection{T7: Motion-Conditioned Generation}
\vspace{-0.1cm}

\mypara{Task formulation.}
Given an initial frame showing all players in their starting positions, a
\emph{player-removed background video} (the original footage with all players
digitally erased via video inpainting, leaving only the court or pitch and
static elements), and a set of player motion trajectories specified as
time-aligned bounding-box sequences, the model must generate a video in which
players follow the prescribed trajectories while remaining visually,
physically, and temporally coherent. Prior trajectory-conditioned generation~\cite{wang2024boximator,yin2023dragnuwa,ma2024trailblazer,namekata2024sg} typically involves one or two objects in simple scenes; T7 targets multi-agent coordination where 10+ players move simultaneously, interact physically, and occlude one another.

\mypara{Data and construction.}
Each instance consists of: (1) an initial frame, (2) per-player motion trajectory as bounding-box sequences, and (3) a player-removed background video
generated via video inpainting~\cite{gen-omnimatte}. We apply explicit quality filtering to remove instances with unstable tracking, severe occlusion, or visible inpainting artifacts (e.g., residual player silhouettes, texture bleeding), ensuring that generation models operate on clean background inputs.

\mypara{Evaluation.}
We evaluate with two metrics: \emph{Video mIoU}~\cite{gberta_2020_CVPR},
measuring spatiotemporal alignment between player trajectories in generated
and reference videos; and \emph{temporal feature similarity}, comparing
SigLIP~\cite{tschannen2025siglip} features from corresponding player regions
across frames to assess visual consistency.

\mypara{Baselines and findings.}
Our reference method fine-tunes Wan~2.1~\cite{wan2025} on \name{} data, extending it to accept structured input conditions (initial frame, bounding boxes, trajectories, player-removed background). We additionally evaluate ATI~\cite{wang2025ati} and MagicMotion~\cite{li2025magicmotion} off-the-shelf. On basketball, our model achieves Video mIoU of 0.513 vs.\ 0.466 (MagicMotion) and 0.397 (ATI), with larger gains on soccer (0.611 vs.\ 0.544 and 0.402). Temporal feature similarity follows the same trend (basketball: 0.787 vs.\ 0.725/0.617; soccer: 0.804 vs.\ 0.708/0.507). Yet even our best model's 0.513 mIoU means roughly half of generated player positions deviate significantly from prescribed trajectories, indicating
that reliable multi-agent motion control remains far from solved.

\subsubsection{T8: Goal-Conditioned Action Generation}

\mypara{Task formulation.}
Given an initial frame, a player-removed background video, and a textual
instruction specifying target player(s), spatial constraints (start and end
bounding boxes), and a desired action outcome (e.g., a rebound, a contested
layup), the model must generate a video in which the specified players execute
a coherent action sequence that achieves the described objective. Unlike T7, which prescribes exact trajectories, T8 requires the model to \emph{plan} intermediate actions to achieve a high-level goal under explicit spatial constraints, requiring implicit understanding of environment dynamics and goal-directed reasoning, going beyond open-ended text-conditioned generation~\cite{blattmann2023videoldm,guo2024animatediff,ho2022video,singer2022makeavideo}.

\mypara{Data and construction.}
We pair curated basketball video clips with structured goal specifications derived from annotated actions, covering diverse goal-conditioned behaviors including completing plays at designated locations, executing specific moves, and interaction-aware scenarios.

\mypara{Evaluation.}
We evaluate with three complementary metrics: \emph{mIoU} on the final frame,
measuring bounding-box overlap between generated and target player positions; \emph{feature
similarity} on the final frame, assessing visual fidelity of the realized
outcome; and \emph{goal accuracy} via a fine-tuned video-language QA model
evaluating whether the generated video achieves the specified objective.

\mypara{Baselines and findings.}
We adapt the Wan-based framework from T7 to the goal-conditioned setting, replacing trajectory inputs
with textual goal specifications and spatial endpoint constraints.
The model achieves final-frame mIoU of 0.344 vs.\ 0.129 (MagicMotion) and 0.047 (ATI), feature similarity of 0.468 vs.\ 0.169 (MagicMotion) and 0.067 (ATI), and goal accuracy of 50.2\% vs.\ 31.4\% (MagicMotion) and 40.5\% (ATI). These results highlight a fundamental gap between trajectory-following (T7) and goal-directed video generation (T8).

\begin{finding}{pillaramber}{Pillar 3 Takeaway (Strategic Simulation).}
Even fine-tuned video generation models cannot reliably coordinate multi-agent motion, with half of generated players deviating from prescribed trajectories. Performance degrades further when models must plan actions toward a goal rather than follow trajectories.
\end{finding}

\vspace{-0.3cm}
\subsection{Pillar 4: Agentic Synthesis (T9)}
\label{sec:pillar4}
\vspace{-0.1cm}



\begin{figure*}[t]
  \centering
\includegraphics[width=0.88\textwidth]{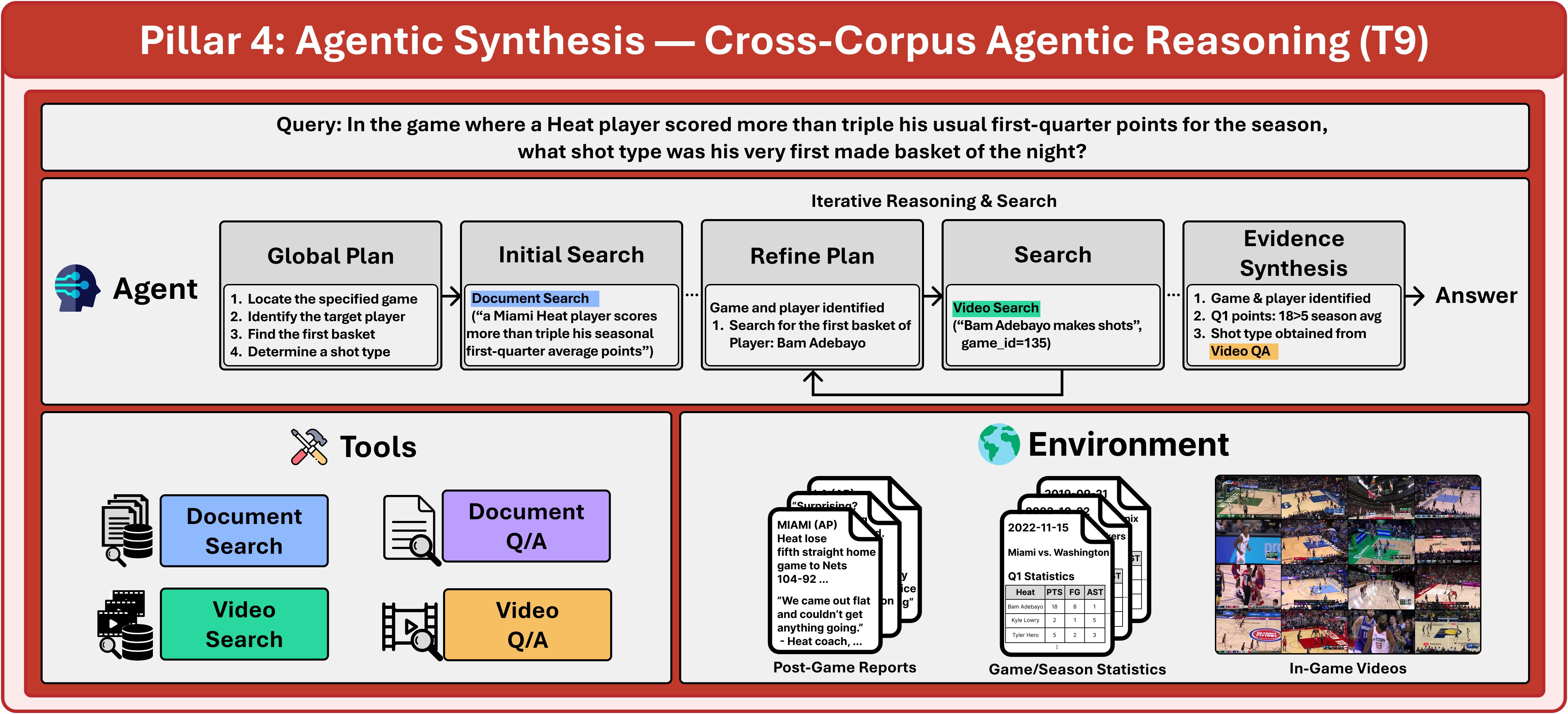}
\vspace{-0.2cm}
\caption{\textbf{Overview of Pillar 4: Agentic Synthesis.} This pillar evaluates the ability to autonomously gather and integrate multimodal evidence through a single task: cross-corpus agentic reasoning (T9), where the agent plans and executes tool-assisted search across large-scale heterogeneous sources to answer complex strategic queries.\vspace{-0.5cm}}
  \label{fig:pillar4_examples}
\end{figure*}

\subsubsection{T9: Cross-Corpus Agentic Reasoning}

\mypara{Task formulation.}
Given a complex natural-language query, the model must plan a multi-step retrieval strategy, gather evidence from heterogeneous sources (video clips, game reports, statistical records), and reason over it to produce a final answer (Figure~\ref{fig:pillar4_examples}). The agent is equipped with search and QA tools over a document database (post-game reports, game- and season-level statistics) and a video database (footage segmented into 10--15s clips). While tool-augmented reasoning has been explored in text~\cite{qin2023toolllm,li2023api,mialon2023gaia,zhou2024webarena} and single video settings~\cite{timesearch-r,yang2025longvt,thinkingwithvideos},  T9 extends it to multimodal evidence at corpus scale, requiring the agent to integrate evidence across modalities through complex reasoning patterns (looping, backtracking, conditional branching, numerical aggregation) over ${\sim}1.8$M clips and ${\sim}$33K documents across three sports.
T9 adopts the hard-to-find but easy-to-verify principle from prior agentic search work~\cite{wei2025browsecompsimplechallengingbenchmark}. Each question begins with seed facts (a specific play, score, or event attribute) and adds multi-hop narrative constraints that uniquely identify the relevant event in the corpus. The answer is a short factual item such as a player number, shot placement, or score. This makes brute-force lookup across 7{,}430 games impractical and forces the agent to search, while the short-answer format supports reliable correctness judgments.

\mypara{Data and construction.}
The corpus covers 7{,}430 basketball, hockey, and soccer games, with 26{,}448 statistical documents, 6{,}859 game reports, and ${\sim}1.8$M video clips (${\sim}5{,}670$ broadcast hours). Questions are constructed to require evidence from multiple sources rather than any single modality alone. The final evaluation set contains 1{,}000 questions, balanced across sports.


\mypara{Evaluation.}
We report accuracy, using an LLM judge (GPT-5.2) to compare the agent's response to the ground-truth answer.



\mypara{Baselines and findings.}
Even the strongest model (GPT-5.2) achieves only 4.6\% accuracy across the three sports, with performance dropping further for smaller open-source models (Qwen3-Omni-30B~\cite{xu2025qwen3}: 2.1\%). Tool-use patterns reveal the same gap: GPT-5.2 issues roughly 21 tool calls per question, while smaller Qwen models make only 3--8 and often terminate early with insufficient evidence. Together, these results show that current agents cannot reliably plan retrieval, gather evidence, and reason over a corpus at this scale.

\begin{finding}{pillarcrimson}{Pillar 4 Takeaway (Agentic Synthesis).}
Agentic synthesis is the hardest capability in \name{}, with even the strongest model reaching only 4.6\%. 
\end{finding}

\vspace{-0.3cm}
\section{Cross-Task Analysis}
\label{sec:cross_task}
\vspace{-0.2cm}

This section analyzes per-task results to characterize
the overall trends in performance from perception through
agentic synthesis. 


\vspace{-0.3cm}
\subsection{The Performance Cliff}
\label{sec:cliff}
\vspace{-0.1cm}


Figure~\ref{fig:capability_cliff} plots the best task score within each pillar, each normalized to a 0--100\% range. The dashed line marks the drop from the strongest perception result to agentic synthesis, with reasoning and simulation in between. The performance drop is consistent across models within each pillar, and gains from task-specific fine-tuning at the perception level do not carry to higher pillars. This suggests that current systems perceive dynamic multi-agent environments far more accurately than they can reason about, simulate, or plan within them.


\vspace{-0.3cm}
\subsection{Oracle Experiments}
\label{sec:oracles}
\vspace{-0.1cm}

\begin{figure}[t]
  \centering
  \includegraphics[width=\linewidth]{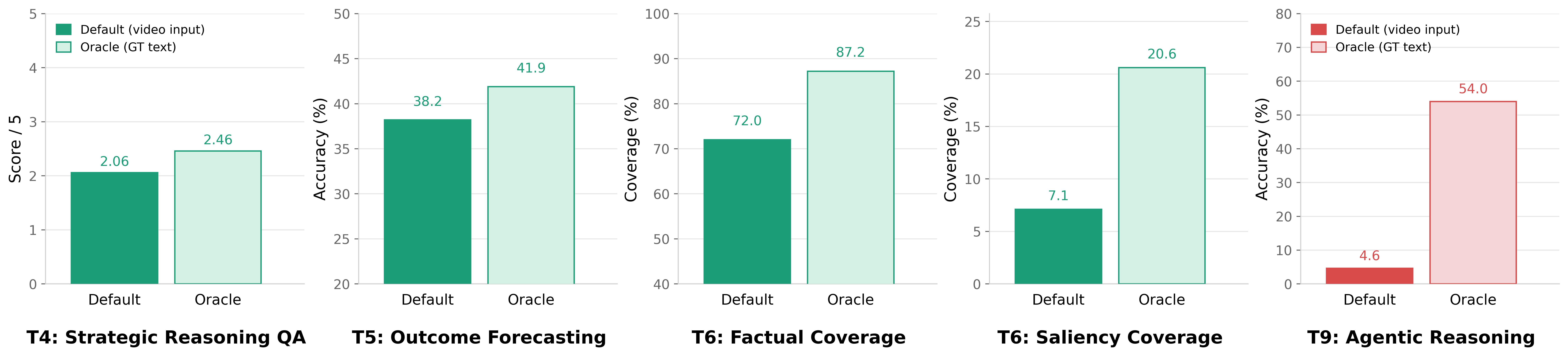}
  \vspace{-0.6cm}

\caption{\textbf{Oracle performance on reasoning and agentic tasks (T4, T5, T6, T9).} The oracle variant replaces video with ground-truth textual descriptions of game events. All tasks use GPT-5.2, except for T6 which uses GPT-5. Gains are small on T4 and T5, moderate on T6, and largest on T9, indicating that strategic reasoning, forecasting, saliency judgment, and multi-step planning remain distinct bottlenecks.\vspace{-0.5cm}}
  \label{fig:oracle}
\end{figure}

To test whether perception is the primary bottleneck in reasoning and agentic tasks, we replace video input with ground-truth textual descriptions of game events derived from play-by-play logs, evaluating the same model in default mode (video) and oracle mode (text) (Figure~\ref{fig:oracle}). The effect varies sharply across tasks. Oracle access raises T9 accuracy from 4.6\% to 54.0\% and T6 factual accuracy from 71.99\% to 87.19\%, with smaller improvements on T4 (2.06 to 2.46 on the 0--5 score), T5 (38.2\% to 41.9\%), and T6 saliency (7.1\% to 20.6\%). These results suggest that no single capability accounts for the gap: strategic reasoning (T4), forecasting (T5), saliency judgment (T6), and multi-step planning (T9) each limit performance independently.


\suppressfloats[t]
\begin{figure}[t]
  \centering
  \includegraphics[width=\linewidth]{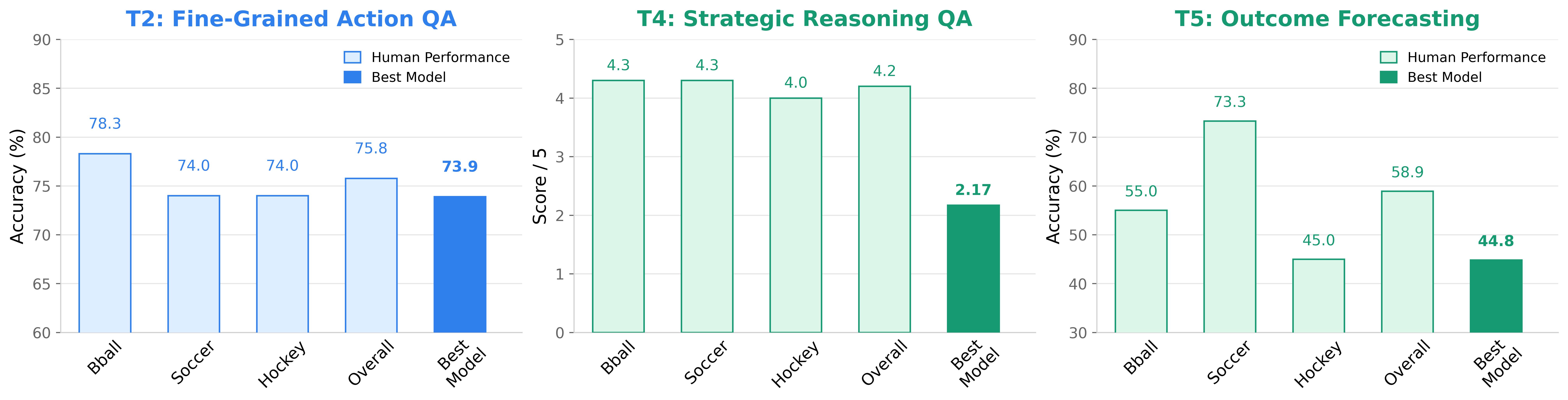}
  \vspace{-0.6cm}

  \caption{\textbf{Human--model comparison on T2, T4, and T5.} Bars show per-sport and overall human performance alongside the best model. T2 and T5 use accuracy (\%). T4 uses the open-ended 0--5 score. Models nearly match humans on perception but trail them substantially on strategic reasoning and forecasting.\vspace{-0.5cm}}
  \label{fig:human_studies}
\end{figure}

\vspace{-0.3cm}
\subsection{Human Studies}
\label{sec:humans}
\vspace{-0.1cm}

To establish that these tasks are solvable and to measure the
human-model gap, we run human studies on T2 (perception), T4
(strategic reasoning), and T5 (forecasting), with participants who
have 5--10 or more years of experience in their sport (Figure~\ref{fig:human_studies}).
Models nearly match humans on perception (T2: 75.8\%
vs.\ 73.9\%) but trail substantially on strategic reasoning (T4:
4.2/5 vs.\ 2.17/5) and forecasting (T5: 58.9\% vs.\ 44.8\%). Humans also know when they are
uncertain: their accuracy rises from 30\% to 90\% across confidence
levels on T2 and from 50\% to 100\% on T5, while GPT-5.2 reports
similar confidence on correct and incorrect answers, producing a
28-point gap between average confidence and accuracy. Human performance establishes that these tasks are
solvable and that the gap reflects current model limitations rather
than task difficulty.

\vspace{-0.3cm}
\section{Conclusion}
\label{sec:conclusion}
\vspace{-0.2cm}

We introduced \name{}, the first large-scale benchmark for strategic
video intelligence in real-world multi-agent video. Across 9 tasks
and four pillars, models perform competently on perception but
degrade substantially on higher-level reasoning tasks. Even given
perfect visual information on our agentic task, the strongest models
reach only 54\%, showing that the challenge extends beyond perception
to reasoning, planning, and evidence integration.

\mypara{Scope and limitations.}
\name{} is a team-sports microworld---a controlled proxy for real
multi-agent video, not a claim of cross-domain generalization.
Sports-specific properties such as broadcast conventions and fixed
rules may not transfer beyond sports. Several tasks rely on LLM
judges, which we validate via human-agreement checks, though some
bias may persist.

\mypara{Future directions.}
The gaps revealed by \name{} point to three directions: video models
that ground long-form reasoning in visual evidence (T4--T6),
generative models with explicit multi-agent dynamics for
goal-directed action (T7--T8), and multimodal agents that plan and
reason across corpora at scale (T9).

\subsubsection*{Acknowledgements} 
This work was supported by Laboratory for Analytic Sciences via NC State University, ONR Award N00014-23-1-2356, Sony Focused Research award, NSF CAREER Award 2541848, Northeastern University startup funds, and the President Joseph E. Aoun Chair.


\bibliographystyle{splncs04}
\bibliography{camera_ready_refs}

\end{document}